\def\paperlanguage{}
\pdfoutput=1 % for arxiv

\documentclass[letterpaper, 10 pt, conference]{ieeeconf}  % Comment this line out if you need a4paper

\usepackage{bm}
\usepackage{cite}
\include{preamble}

\newcommand{\ctext}[1]{\raise0.2ex\hbox{\textcircled{\scriptsize{#1}}}}

\IEEEoverridecommandlockouts                              % This command is only needed if
\title{\LARGE \textbf
  {
    \switchlanguage%
    {%
      Robotic Constrained Imitation Learning for the Peg Transfer Task\\in Fundamentals of Laparoscopic Surgery
    }%
    {%
      腹腔鏡手術の基本技能訓練におけるペグ移動タスクに向けたロボットの制約付き模倣学習
    }%
  }
}

\author{Kento Kawaharazuka$^{1}$, Kei Okada$^{1}$, and Masayuki Inaba$^{1}$% <-this % stops a space
  \thanks{$^{1}$ The authors are with the Department of Mechano-Informatics, Graduate School of Information Science and Technology, The University of Tokyo, 7-3-1 Hongo, Bunkyo-ku, Tokyo, 113-8656, Japan.
    {\texttt\small [kawaharazuka, k-okada, inaba]@jsk.t.u-tokyo.ac.jp}
  }
}
\begin{document}

\maketitle
\thispagestyle{empty}
\pagestyle{empty}

%%%%%%%%%%%%%%%%%%%%%%%%%%%%%%%%%%%%%%%%%%%%%%%%%%%%%%%%%%%%%%%%%%%%%%%%%%%%%%%%
\begin{abstract}
  \switchlanguage%
  {%
    In this study, we present an implementation strategy for a robot that performs peg transfer tasks in Fundamentals of Laparoscopic Surgery (FLS) via imitation learning, aimed at the development of an autonomous robot for laparoscopic surgery.
    Robotic laparoscopic surgery presents two main challenges: (1) the need to manipulate forceps using ports established on the body surface as fulcrums, and (2) difficulty in perceiving depth information when working with a monocular camera that displays its images on a monitor.
    Especially, regarding issue (2), most prior research has assumed the availability of depth images or models of a target to be operated on.
    Therefore, in this study, we achieve more accurate imitation learning with only monocular images by extracting motion constraints from one exemplary motion of skilled operators, collecting data based on these constraints, and conducting imitation learning based on the collected data.
    We implemented an overall system using two Franka Emika Panda Robot Arms and validated its effectiveness.
  }%
  {%
    本研究では腹腔鏡手術を行うロボットの開発に向けた, Fundamentals of Laparoscopic Surgery (FLS)におけるペグ移動を模倣学習により行うロボットの実装戦略について紹介する.
    ロボットによる腹腔鏡手術には(1)体表に開けたポートを支点として鉗子を動かす必要がある, (2)単眼カメラの映像をモニタに映して作業するため奥行き方向を認識しにくい, という問題点がある.
    特に(2)について, これまでの研究のほとんどが, 深度画像や対象のモデルが得られることが前提である.
    そこで本研究では, 単眼の画像のみから熟練者の動作に基づく動作制約の抽出とこれに基づくデータ収集, 集めたデータに基づく模倣学習を行うことで, より高い精度での模倣学習を実現する.
    2台のFranka Emika Panda Robot Armを用いて全体システムを実装し, その有効性を確かめた.
  }%
\end{abstract}

\section{INTRODUCTION}\label{sec:introduction}
\begin{figure}[t]
  \centering
  \includegraphics[width=0.9\columnwidth]{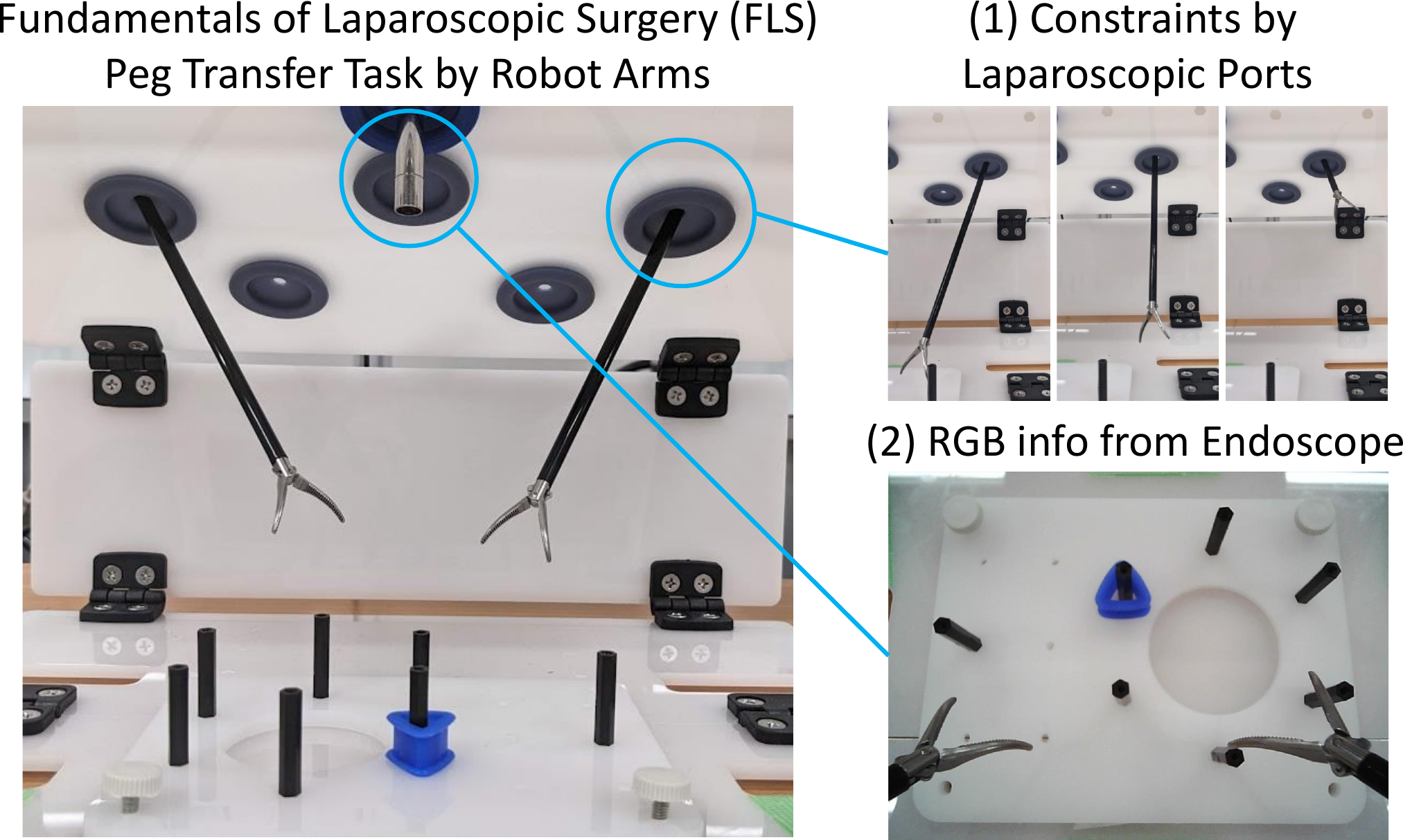}
  \vspace{-1.0ex}
  \caption{The concept of this study: regarding peg transfer tasks in Fundamentals of Laparoscopic Surgery (FLS) for robots using imitation learning, we handle two main problems: (1) the forceps are constrained by laparoscopic ports and (2) only RGB information can be obtained from a monocular endoscope.}
  \vspace{-3.0ex}
  \label{figure:concept}
\end{figure}

\switchlanguage%
{%
  Laparoscopy is a minimally invasive surgical procedure with less scarring and less postoperative pain compared to laparotomy \cite{clarke1972laparoscopy}, which is highly intriguing as an example of a non-repetitive task required in scientific experiments.
  In this study, we aim to develop a robot that can perform this laparoscopic surgery autonomously via imitation learning.
  The nature of laparoscopic surgery, in which an endoscope and forceps are inserted into ports on the body surface, imposes several constraints on the movements of the operator and the robot.
  They are roughly divided into problem (1) and (2) as shown in \figref{figure:concept}: (1) it is necessary to move the forceps using the laparoscopic ports as fulcrums so as not to put load on the ports, and (2) it is difficult to perceive the depth information because images captured by the endoscope are being viewed through a monitor.
  % Especially, regarding (2), most prior research has assumed the presence of depth images or models of a target intended for manipulation.
  In this study, we introduce an implementation strategy of a laparoscopic surgical robot that solves these two problems.
  As a target task, we handle the peg transfer task in Fundamentals of Laparoscopic Surgery (FLS), which is known as an effective training for laparoscopic surgery \cite{peters2004development}.
  While FLS significantly simplifies the actual environment, it captures the overall characteristics of laparoscopic surgery, and we believe that the insights gained here can be effectively utilized.
}%
{%
  体内に挿入した内視鏡によりモニタ越しに映像を確認し, 同様に体内に挿入した鉗子を操作して行う腹腔内手術(laparoscopy)は, 一般的な開腹手術(laparotomy)と比較して, 小さな傷跡で術後の痛みも少ない低侵襲な手術方法である\cite{clarke1972laparoscopy}.
  laparoscopic surgeryは, scientific experimentで必要とされるnon-repetitive taskの例として非常に興味深い.
  本研究ではこの腹腔鏡手術を自律的に行うロボットを目指す.
  この腹腔鏡手術は体表に開けたポートに内視鏡や鉗子を挿入するという性質上, 人間, そしてロボットの動作にいくつかの制約が課される.
  それらは大きく(1)と(2)の2つに分けられる(\figref{figure:concept}).
  (1) 体表に開けたポートに負荷をかけないよう, そのポートを支点として鉗子を動かす必要がある.
  (2) 内視鏡により捉えた映像をモニタ越しに確認するため, 奥行き方向の認識が難しい.
  本研究では主にこの2つの問題点を解決する腹腔鏡手術ロボットの実装戦略について紹介する.
  腹腔鏡手術の効果的なトレーニングとして知られるFundamentals of Laparoscopic Surgery (FLS)におけるペグ移動\cite{peters2004development}を題材として扱う.
  FLSは実際の体内環境を大きく単純化しているものの, 腹腔鏡手術の全体的な特性は捉えており, ここで得られた知見は有効に利用できると考えている.
}%

\begin{figure*}[t]
  \centering
  \includegraphics[width=1.9\columnwidth]{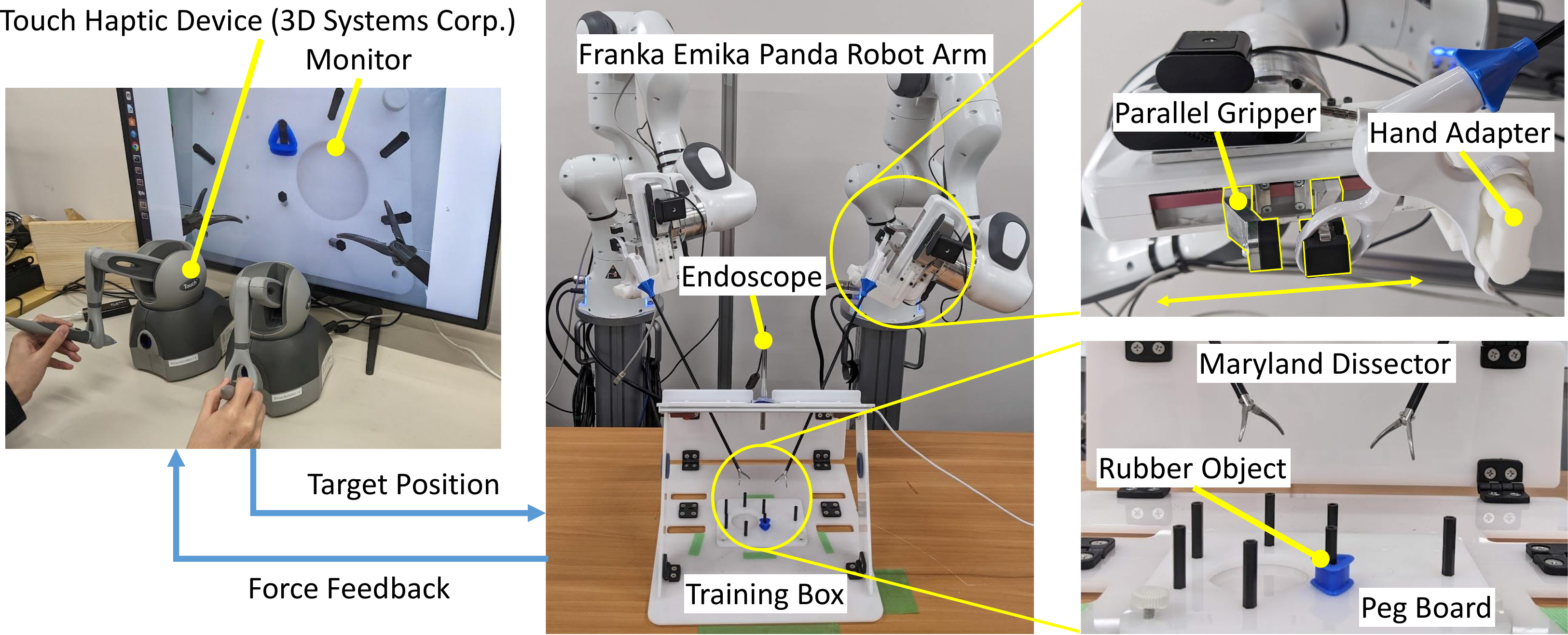}
  \vspace{-1.0ex}
  \caption{The setup for robotic peg transfer tasks in Fundamentals of Laparoscopic Surgery. Franka Emika Panda Robot Arms are operated by Touch Haptic Device (3D Systems Corp.). Maryland Dissectors are controlled via Hand Adapter and Parallel Gripper to transfer Rubber Object on Peg Board.}
  \vspace{-1.0ex}
  \label{figure:fls-setup}
\end{figure*}

\switchlanguage%
{%
  We introduce several related works.
  \cite{preda2015surgical} has developed a motion planning method for a suturing surgical robot based on RRT-connect \cite{kuffner2000rrtconnect} and multiple primitive motions.
  \cite{sen2016suturing} has discussed a needle guide mechanism and its trajectory optimization for a suturing surgical robot.
  \cite{seita2018surgical} has proposed a calibration method of da Vinci Research Kit (dVRK) \cite{kazanzides2014davinchkit} and performed an autonomous debridgment based on stereo image and edge recognition.
  \cite{meli2021segmentation, rossi2021segmentation} have discussed surgical action segmentation, and \cite{rossi2021segmentation} has performed autonomous task execution using the segmentation and model predictive control.
  In recent years, the application of deep learning to surgical robots has been expanding \cite{blanco2021surgery}.
  \cite{saeidi2022laparoscopic} has successfully performed a series of laparoscopic surgeries using deep learning-based tissue motion tracking and motion planning.
  \cite{thananjeyan2017cutting} has proposed a method for surgical pattern cutting by applying tension to the gauze based on reinforcement learning.
  \cite{kim2021surgery} has proposed an eye surgery method using imitation learning.
  On the other hand, most of them are based on the assumption that a depth image or a model of the target is available.
  Of course, there are stereoscopic endoscopes, but the basic technique is still based on a monocular image projected on a monitor.
  % Moreover, there have been no attempts to perform laparoscopic surgery or its training by autonomous operation of robots.
  The purpose of this study is to address the problems unique to laparoscopic surgery, and to improve the autonomy of the robot based on imitation learning.
  In particular, we propose and discuss a constrained imitation learning, which solves the difficulty in recognizing depth direction due to monocular images by generating constraints based on a single exemplary demonstration.
  This is a new approach that does not devise a learning method but devises its data collection based on the extracted constraints, and we show that this approach is useful for laparoscopic surgery.
  Our contributions are as follows.
  \begin{itemize}
    \item A constrained imitation learning method that collects data with extracted constraints from a single exemplary demonstration and trains a predictive model.
    \item An overall setup of a laparoscopic surgery robot system and control architecture that can perform the peg transfer task in FLS.
  \end{itemize}

  This study is organized as follows.
  In \secref{sec:proposed}, we describe the task to be performed, the robot configuration, the experimental environment, inverse kinematics considering constraints on body surface ports, and imitation learning using the data collected considering constraints generated from one exemplary demonstration.
  In \secref{sec:experiment}, we describe experiments of constraint generation from one exemplary demonstration, data collection for the peg transfer task with the generated constraints, and the peg transfer experiment based on imitation learning.
  In \secref{sec:discussion}, we discuss the experimental results and limitations of this study, and conclude in \secref{sec:conclusion}.
}%
{%
  いくつかの先行研究について紹介する.
  \cite{preda2015surgical}はRRT-connect \cite{kuffner2000rrtconnect}と複数のプリミティブ動作に基づく縫合手術ロボットの動作計画を開発した.
  \cite{sen2016suturing}は縫合手術ロボットに向けた針のガイドメカニズムとその軌跡の最適化について議論している.
  \cite{seita2018surgical}はda Vinci Research Kit (dVRK) \cite{kazanzides2014davinchkit}のキャリブレーション手法を提案し, ステレオ画像とエッジ認識に基づく自律的なdebridgementを行った.
  \cite{meli2021segmentation, rossi2021segmentation}は手術動作のセグメンテーションについて議論しており, 特に\cite{rossi2021segmentation}はこれとMPCを用いた自律的なタスク実行を行った.
  近年は手術ロボットに向けた深層学習の適用も広がりつつある\cite{blanco2021surgery}.
  \cite{saeidi2022laparoscopic}は深層学習に基づくtissue motion trackingと動作計画により一連の腹腔鏡手術に成功している.
  \cite{thananjeyan2017cutting}は強化学習に基づきガーゼにテンションを加えることでsurgical pattern cuttingを行う手法を提案している.
  また, \cite{kim2021surgery}は模倣学習を用いたeye surgeryについて提案している.
  一方で, 大抵は深度画像や対象のモデルが得られることが前提である.
  もちろんstereoscopic endoscopeも存在するが, 未だに基本はモニタに単眼画像を映すことによる術式である.
  % また, これまで腹腔鏡手術やその訓練をロボットの自律動作によって行う試みはない.
  本研究は上記で述べた腹腔鏡手術特有の問題点を扱い, 模倣学習に基づきロボットの自律性を向上させることを目的としている.
  特に, 単眼画像のため奥行き方向の認識が難しい問題を, 熟練者動作に基づく制約生成により解決する制約付き模倣学習について提案・議論する.
  これは, 学習方法を工夫するのではなく, そのデータ取得を制約に基づき工夫するという新しいアプローチであり, これが腹腔鏡手術に有用であることを示す.

  本研究の構成は以下である.
  \secref{sec:proposed}では, 行うタスクとロボット構成や実験環境, 体表ポート制約付き逆運動学, 熟練者動作における制約つき模倣学習について述べる.
  \secref{sec:experiment}では, 熟練者動作に基づくペグ移動の制約生成実験とデータ収集, 模倣学習に基づくペグ移動実験について順に述べる.
  \secref{sec:discussion}では, 本研究における実験結果といくつかの限界について考察し, \secref{sec:conclusion}で結論を述べる.
}%

\section{Fundamentals of Laparoscopic Surgery for Robots with Constrained Imitation Learning} \label{sec:proposed}

\subsection{Fundamentals of Laparoscopic Surgery for Robots} \label{subsec:fls-setup}
\switchlanguage%
{%
  The overall setup of this study is shown in \figref{figure:fls-setup}.
  The robots used in this study are two Franka Emika Panda Robot Arms.
  A FLS box training kit is placed in front of the robots.
  A Maryland Dissector is attached to each hand of the robot via a Hand Adapter.
  A parallel gripper is used to open and close the forceps, and the rotation of forceps in the long axis direction is fixed.
  The forceps pass through the ports of the box training kit, and rubber object is operated on peg board.
  Similarly, an endoscope passes through the port and projects the image of the entire peg board on a monitor.
  Although a peg transfer task is originally performed by moving each rubber object stuck to the six pegs, as a simpler setup, the task in this study is to insert a rubber object stuck to one of the three pegs into the single left peg on the right side from the robot's viewpoint.
  Two Touch Haptic Devices (3D Systems Corp.) are prepared in front of the monitor for demonstration by experts.
  The haptic devices can send the target positions of the tips of the forceps to the robot, and at the same time, the forces received by the robot can be reflected to the haptic devices.
  Note that bilateral control is not performed due to the limitation of the body surface ports and the significant difference in the structure of Touch Haptic Device and Panda Robot Arm.
}%
{%
  本研究の全体的なセットアップを示す(\figref{figure:fls-setup}).
  用いたロボットは2台のFranka Emika Panda Robot Armである.
  FLSのボックストレーニングキットを前方に配置する.
  それぞれのロボットのハンドにはMaryland DissectorがHand Adapterを介して装着されている.
  Parallel Gripperによって鉗子の開閉を行うが, 長軸方向の回転については固定している.
  鉗子はボックストレーニングキットのポートを通り, Peg Board上でRubber Objectを操作する.
  同様に内視鏡がポートを通してPeg Board全体の映像をモニタに映す.
  ペグ移動は本来は6つのペグに刺さったrubber objectそれぞれを移動させるが, 本研究ではより簡易なセットアップとして, ロボットから見て右側の3つのPegいずれかに刺さったRubber Objectを左の一本のPegに挿入するタスクを行う.
  エキスパートによるデモンストレーションを行うため, 操作デバイスとしてTouch Haptic Device (3D Systems Corp.)を2台, モニタの前に用意する.
  この操作デバイスから道具の先端位置をロボットに送ることができると同時に, ロボットが受けた力を操作デバイス側に反映させることができる.
  なお, 体表ポートの制約やTouch Haptic DeviceとPanda Robot Armの構造の大きな違いからバイラテラル制御は行っていない.
}%

\begin{figure}[t]
  \centering
  \includegraphics[width=0.9\columnwidth]{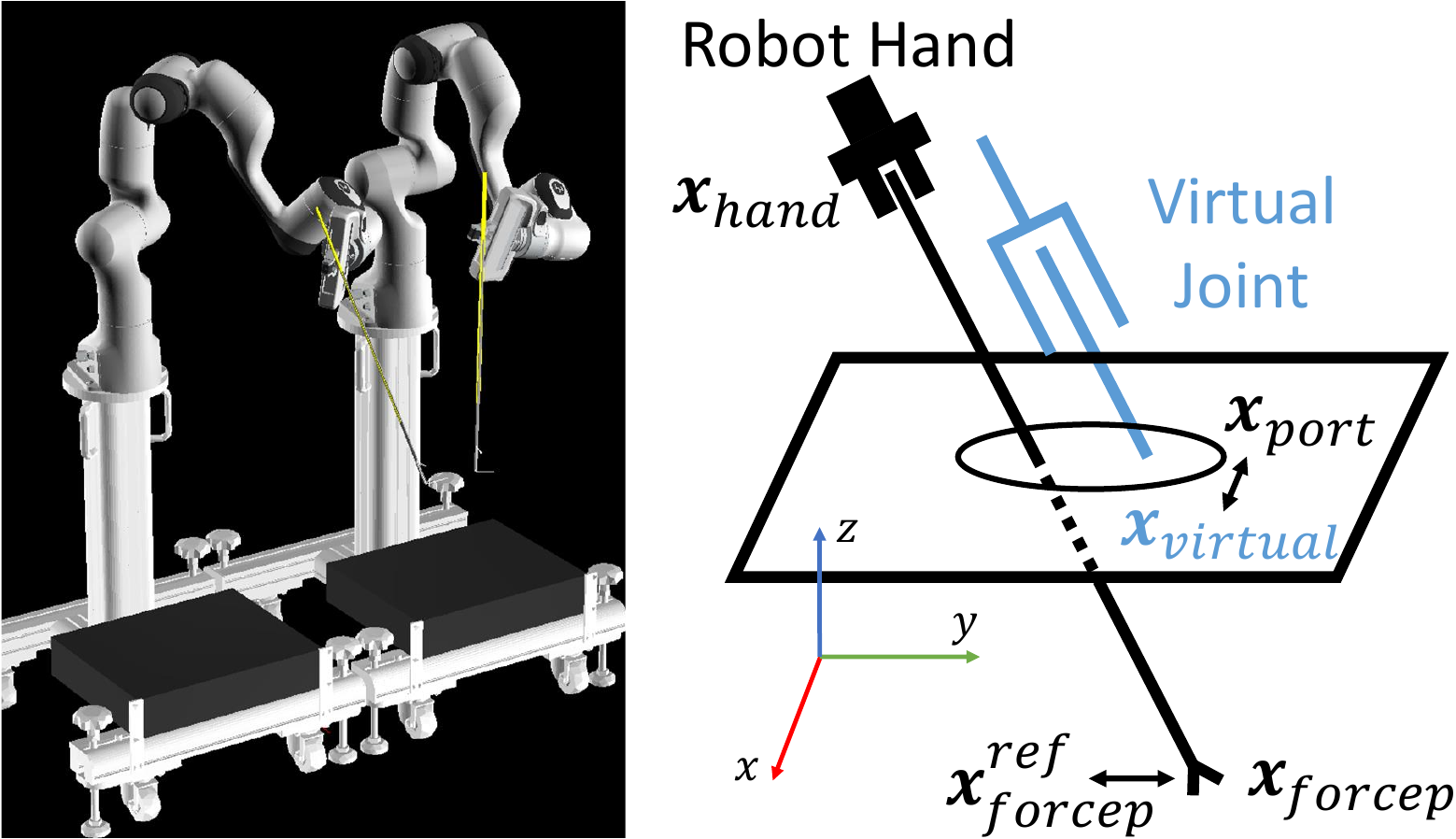}
  \vspace{-1.0ex}
  \caption{The configuration to control forceps considering the constraint by a laparoscopic port. The left figure shows a geometric model of robot arms, and the right figure shows the kinematics of forceps with a virtual linear joint from the tip of the hand that overlaps with the long axis of the forceps.}
  \vspace{-3.0ex}
  \label{figure:fls-ik}
\end{figure}

\subsection{Constrained Inverse Kinematics for Laparoscopic Surgery} \label{subsec:fls-ik}
\switchlanguage%
{%
  In laparoscopic surgery, the robot needs to solve inverse kinematics with body surface port constraints.
  Although a human can sense the constraint appropriately by sensing the force received from the port, it is difficult to do so with the force resolution of Panda Robot Arm and Touch Haptic Device.
  Therefore, we determine the position of the body surface port in advance and apply the constraint to motion generation.
  \figref{figure:fls-ik} shows a geometric model of the robot and its constraints.
  Let $\bm{x}_{hand}$ denote the coordinate of the robot hand, $\bm{x}_{forcep}$ denote the tip of the Maryland Dissector, and $\bm{x}_{port}$ denote the position of the body surface port.
  Here, the forceps not only rotate but also move in the direction of its long axis, for a total of three degrees of freedom.
  In order to take this constraint into account, we add a virtual linear joint from the tip of the hand that overlaps with the long axis of the forceps.
  Let $\bm{x}_{virtual}$ be the tip of this virtual joint.
  That is, given a target value $\bm{x}^{ref}_{forcep}$ of the forcep-tip position from the haptic device, the following inverse kinematics $\textrm{IK}$ is calculated so that $\bm{x}_{forcep}$ becomes close to $\bm{x}^{ref}_{forcep}$ and $\bm{x}_{virtual}$ becomes close to $\bm{x}_{port}$,
  \begin{align}
    \bm{\theta}, \bm{\theta}_{virtual} = \textrm{IK}(\bm{x}_{forcep}, \bm{x}^{ref}_{forcep}, \bm{x}_{virtual}, \bm{x}_{port})
  \end{align}
  where $\bm{\theta}$ is the joint angle of the robot arm and $\bm{\theta}_{virtual}$ is the angle of the virtual linear joint.

  From now on, the right and left hands are represented in forms such as $\bm{x}_{hand-left}$ and $\bm{x}_{forcep-right}$.
  It may also be expressed in forms such as $x_{hand-right}$ and $z^{ref}_{forcep-left}$ by setting $\bm{x}^{T}=\begin{pmatrix}x&y&z\end{pmatrix}$.
  Note that solving this problem can also be achieved through prioritized inverse kinematics or optimization methods.
  % It does not introduce significantly novel concepts but is part of the overall system.
}%
{%
  腹腔鏡手術では体表ポート制約付きの逆運動学をロボットは解く必要がある.
  人間であればポートから受ける力を感じて適切に制約を感じ取ることができるが, 本研究のPanda RobotとTouch Haptic Deviceの力の分解能ではそれは難しかった.
  そこで, 予め体表ポートの位置を決め, これにより制約をかけて動作生成を行う.
  \figref{figure:fls-ik}はロボットの幾何モデルとその制約を示している.
  ロボットのハンドの座標を$\bm{x}_{hand}$, Maryland Dissectorの先端を$\bm{x}_{forcep}$, 体表ポートを位置を$\bm{x}_{port}$とする.
  ここで, 鉗子は回転だけではなく鉗子の長軸方向の計3自由度を動くことになる.
  この制約を考慮するため, ハンド先端から, 鉗子の長軸方向に重なる仮想の直動ジョイントを追加した.
  この仮想ジョイントの先端を$\bm{x}_{virtual}$とする.
  つまり, 操作デバイスから道具先端位置の指令値$\bm{x}^{ref}_{forcep}$が与えられたとき, $\bm{x}_{forcep}$を$\bm{x}^{ref}_{forcep}$に, $\bm{x}_{virtual}$を$\bm{x}_{port}$に近づけるように, 以下の逆運動学$\textrm{IK}$を解く.
  \begin{align}
    \bm{\theta}, \bm{\theta}_{virtual} = \textrm{IK}(\bm{x}_{forcep}, \bm{x}^{ref}_{forcep}, \bm{x}_{virtual}, \bm{x}_{port})
  \end{align}
  なお, ここで$\bm{\theta}$はロボットの関節角度, $\bm{\theta}_{virtual}$は仮想ジョイントの関節角度を指す.

  今後右手左手を表す場合は, $\bm{x}_{hand-left}$, $\bm{x}_{forcep-right}$の形で表現する.
  また, $\bm{x}^{T}=\begin{pmatrix}x&y&z\end{pmatrix}$とし, $x_{hand-right}$や$z_{forcep-left}$の形でも表現することがある.
  なお, これについては優先度付き逆運動学や最適化問題でも解くことが可能であり, 大きく新規な内容ではなく, あくまで全体システムの一部である.
}%

\begin{figure}[t]
  \centering
  \includegraphics[width=0.85\columnwidth]{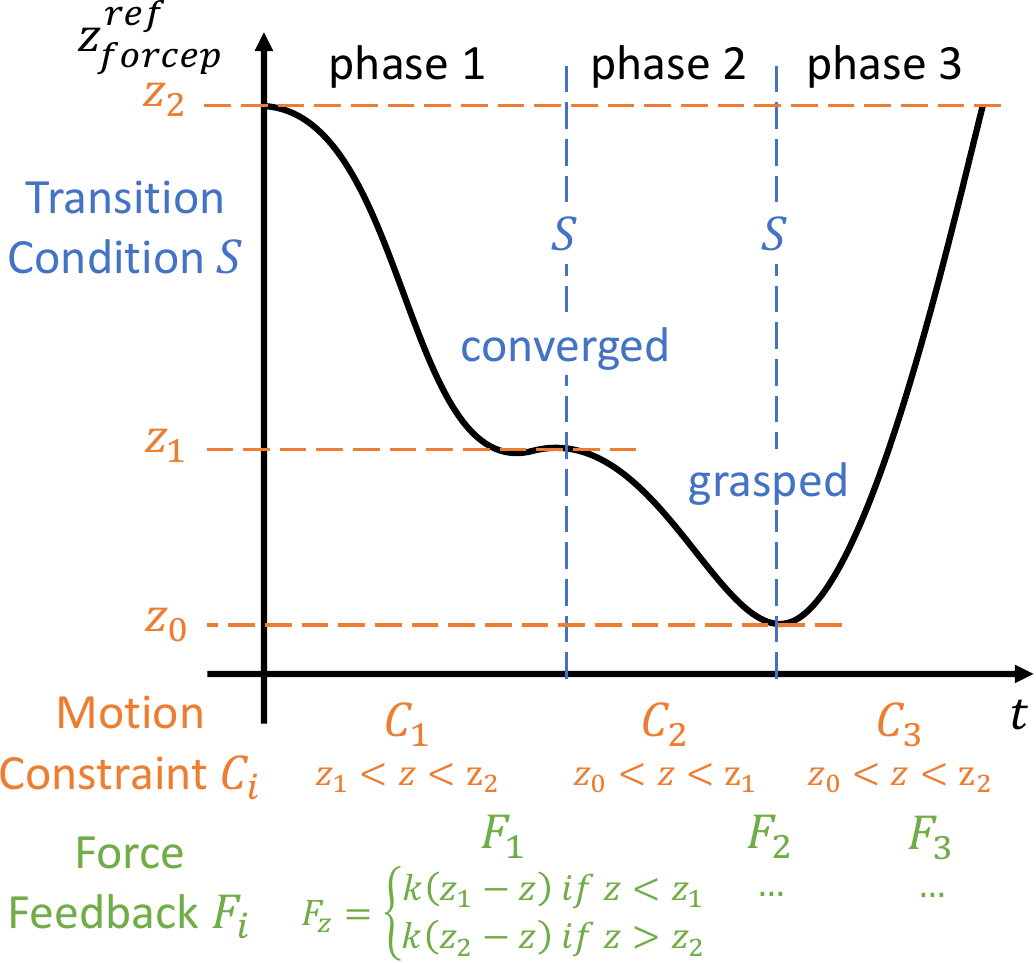}
  \vspace{-1.0ex}
  \caption{Trajectory example of target forcep-tip position $z^{ref}_{forcep}$. The transition condition $S$ divides phases of the demonstration. Motion constraint $C_{i}$ and force feedback function $F_{i}$ are generated from the motion in each phase $i$.}
  \vspace{-3.0ex}
  \label{figure:fls-learning}
\end{figure}

\begin{figure*}[t]
  \centering
  \includegraphics[width=1.95\columnwidth]{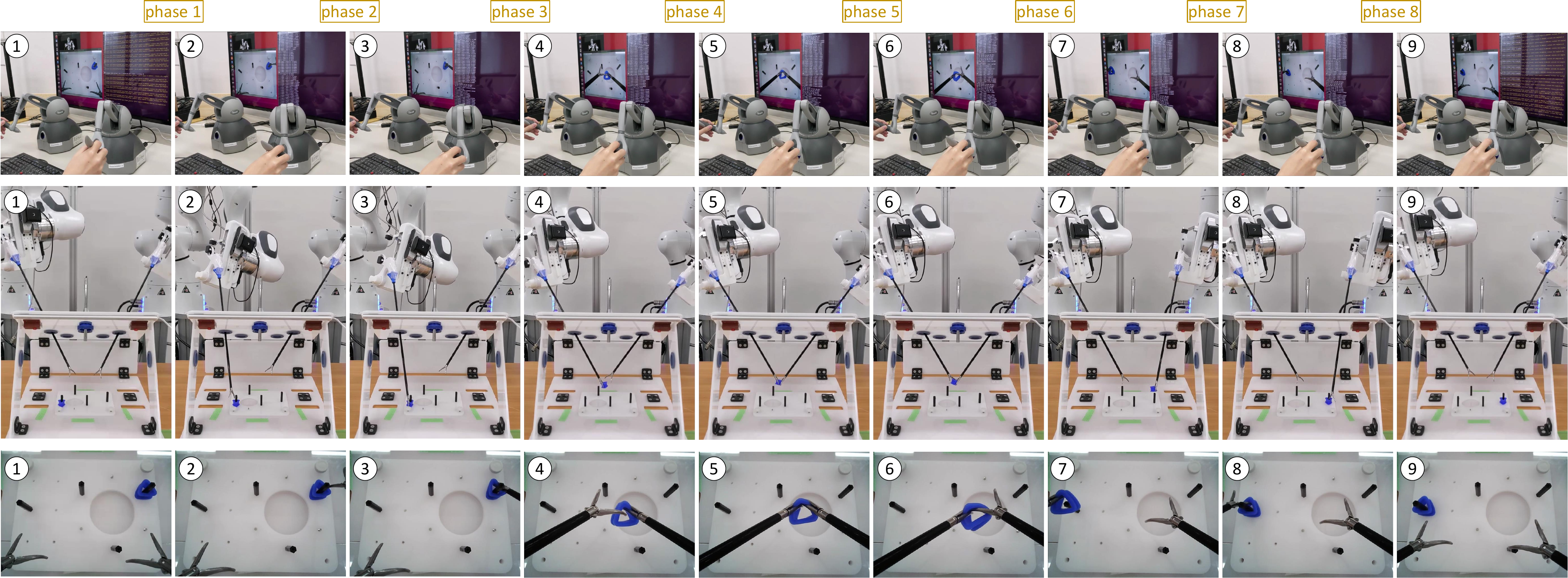}
  \vspace{-1.0ex}
  \caption{One exemplary demonstration of peg transfer. The upper figures show the human teaching with haptic devices, the middle figures show the motion of robot arms, and the lower figures show the endoscopic image. Each figure represents the image when there is a phase transition.}
  \vspace{-3.0ex}
  \label{figure:complete-exp}
\end{figure*}

\subsection{Constrained Data Collection and Imitation Learning} \label{subsec:fls-learning}
\switchlanguage%
{%
  In order to generate robot motions by imitation learning, we collect human teaching data using haptic devices.
  However, it is challenging to manipulate the forceps successfully because it is difficult to obtain the depth information.
  Therefore, we extract constraints on the movement of the forceps from the trajectory of a single slow and accurate exemplary demonstration, and incorporate them to improve the quality of the teaching data and the accuracy of imitation learning.
  The procedure is as follows.
  \begin{enumerate}
    \renewcommand{\labelenumi}{\arabic{enumi})}
    \item Describe a phase transition condition
    \item Extract motion constraints pertaining to each phase from a single exemplary demonstration
    \item Collect data by human teaching with force feedback based on the constraints
    \item Execute imitation learning based on the collected data
  \end{enumerate}
  Note that this method is not limited to the FLS task, but can be applied to any task by changing the phase transition condition and motion constraint extraction.

  1) Since motion constraint depends on the motion, a long motion must be divided into phases.
  Therefore, we describe the condition $S$ under which the phase transition or the change in motion constraint occurs.
  As an example, let us consider the following motions: aligning the forceps to a position slightly above the rubber object, then lowering the forceps, grasping the object, and lifting it up.
  Here, in order to align the forceps to a position slightly above the object, we can apply a minimum constraint to the $z$ direction after aligning the forceps to the $xy$ direction.
  In the subsequent motion of lowering the forceps, the minimum value of $z$ is lowered without changing the $xy$ position, and the rubber object becomes graspable.
  The transition of $z^{ref}_{forcep}$ is shown in \figref{figure:fls-learning}.
  Here, the phase transition condition can be, for example, the opening and closing of the forceps, or the convergence of the motion.
  That is, if the velocity of $\bm{x}$ remains small for a certain period of time, it is judged that the positioning has converged and moves on to the next phase, and if a grasping motion occurs, it is judged that the gripper is ready for a lifting motion and moves on to the next phase.
  Other phase transition conditions can be described in any form, such as the change in distance between the two forceps, reversal of its velocity, state change in the buttons of haptic devices, and so on.

  2) Assuming that a single exemplary demonstration is obtained, it is divided into $N_{C}$ phases by the condition $S$ in 1).
  From the demonstration in phase $i$, we extract a motion constraint $C_{i}$ in phase $i$ ($1\leq{i}\leq{N}_{C}$).
  Let us consider the same example as in 1) shown in \figref{figure:fls-learning}.
  In this case, the maximum and minimum value constraints are effective.
  When $z_{\{0, 1, 2\}}$ in \figref{figure:fls-learning} is obtained from demonstration, the constraints on $z^{ref}_{forcep}$ in each phase can be described as follows.
  \begin{align}
    C_1: z_1 < z^{ref}_{forcep} < z_2 \label{eq:constrain-example}\\
    C_2: z_0 < z^{ref}_{forcep} < z_1 \\
    C_3: z_0 < z^{ref}_{forcep} < z_2
  \end{align}
  Of course, constraints can be imposed on $x$ and $y$ as well, but given the difficulty in recognizing the depth information, constraints on $z$ are the most appropriate.
  It is possible to extract arbitrary constraints such as the distance between two forceps, the velocity of the forceps, and so on.

  3) We incorporate the constraints obtained in 2) into the human teaching as force feedback to the haptic device.
  For example, if the constraints of \equref{eq:constrain-example} are obtained, the force feedback can be described as follows,
  \begin{align}
    F_{z} =
    \begin{cases}
      k_{p}(z_1-z^{ref}_{forcep}) & \text{if $z^{ref}_{forcep} < z_1$}\\
      k_{p}(z_2-z^{ref}_{forcep}) & \text{if $z^{ref}_{forcep} > z_2$}
    \end{cases}
  \end{align}
  where $F_{z}$ is the feedback force of the haptic device in the $z$ direction and $k_{p}$ is a proportionality constant.
  This corresponds to applying a feedback control that prevents $z$ from deviating from the range of $z_1$ and $z_2$.
  It is possible to incorporate different constraints by force feedback to $F_x$ and $F_y$ in the same way.
  After adding these constraints, data is collected by human teaching.

  4) Imitation learning is performed based on the collected data.
  To consider various motion speeds and styles in the human teaching, we use a learnable network input variable called parametric bias \cite{tani2002parametric, kawaharazuka2021imitation}.
  The following predictive model-type network is trained,
  \begin{align}
    (\bm{s}_{t+1}, \bm{u}_{t+1}) = \bm{f}(\bm{s}_{t}, \bm{u}_{t}, \bm{p}) \label{eq:rnnpb}
  \end{align}
  where $t$ is the current time step, $\bm{s}$ is the sensor state, $\bm{u}$ is the control input, $\bm{p}$ is the parametric bias, and $\bm{f}$ is the prediction model.
  In this study, $\bm{s}$ is $\bm{\xi}$ ($\in \mathcal{R}^{12}$), which is the image information compressed through AutoEncoder \cite{hinton2006reducing}.
  For $\bm{u}$, we use the target values of the left and right forcep-tips $\bm{x}^{ref}_{forcep-\{left, right\}}$ ($\in \mathcal{R}^{6}$) and the opening and closing state of the forceps $h_{\{left, right\}}$ ($\in \{0, 1\}^{2}$).
  These data points are acquired at 10 Hz.

  For one demonstration $k$, we obtain the data $D_k=\{(\bm{s}_1, \bm{u}_{1}), (\bm{s}_2, \bm{u}_2), \cdots, (\bm{s}_{T_{k}}, \bm{u}_{T_{k}})\}$ ($1 \leq k \leq K $, where $K$ is the total number of demonstrations and $T_{k}$ is the number of time steps for the demonstration $k$).
  Then, we create the data $D_{train}=\{(D_1, \bm{p}_1), (D_2, \bm{p}_2), \cdots, (D_{K}, \bm{p}_K)\}$ for training.
  $\bm{p}_k$ ($1 \leq k \leq K$) is parametric bias, a learnable input variable for the demonstration $k$, and the human motion style is self-organized in this space.
  Using $D_{train}$, we train Recurrent Neural Network with Parametric Bias (RNNPB) by simultaneously updating the network weight $\bm{W}$ and each $\bm{p}_{k}$. %, instead of updating only $W$ as in the usual imitation learning.
  The loss function is a mean squared error, and each $\bm{p}_{k}$ is optimized with an initial value of $\bm{0}$.
  The execution of the task is simple: the current sensor state $\bm{s}_{t}$ and the control input $\bm{u}_{t}$ are obtained, $\bm{u}_{t+1}$ is obtained by calculating \equref{eq:rnnpb}, and $\bm{u}_{t+1}$ is repeatedly commanded to the actual robot.
  For $h$, the forceps is closed when the value exceeds 0.5, and opened when the value falls below 0.5.
  It should be noted that during task execution, $\bm{p}$ can be arbitrarily set, allowing for variations in motion style \cite{kawaharazuka2021imitation}.
  % However, for the main focus of this study, which is the comparison of the presence or absence of 1)-3), we set $\bm{p}=\bm{0}$, aiming for an average motion.

  RNNPB consists of 10 layers, which are 4 fully-connected layers, 2 LSTM \cite{hochreiter1997lstm} layers, and 4 fully-connected layers, in order.
  For the number of units, we set \{$N_u+N_s+N_p$, 500, 300, 100, 100 (number of units in LSTM), 100 (number of units in LSTM), 100, 300, 500, $N_u+N_s$\} (note that $N_{\{u, s, p\}}$ is the number of dimensions of $\{\bm{u}, \bm{s}, \bm{p}\}$).
  The activation function is hyperbolic tangent, and the update rule is Adam \cite{kingma2015adam}.
  Regarding image compression, for an RGB image of $128\times96$, convolutional layers of kernel size 3 and stride 2 are applied 5 times, and the number of units is reduced to 1024 and 12 by the fully-connected layers, and then the image is reconstructed by the fully-connected layers and the deconvolutional layers in the same manner.
  Batch normalization \cite{ioffe2015batchnorm} is applied except to the last layer.
  The activation function is ReLU \cite{nair2010relu} except for the last layer, which uses Sigmoid.
  During the training process, random noises are added to the data (improved version of \cite{florence2020imitation}).
  For $\begin{pmatrix}\bm{s}^{T}&\bm{u}^{T}\end{pmatrix}^{T}$, we compute the difference from the previous step and obtain its covariance $\bm{\Sigma}$.
  Random numbers following a multivariate normal distribution with mean $\bm{0}$ and covariance $0.3^{2}\bm{\Sigma}$ is added to the data in each step.
}%
{%
  本研究では模倣学習を用いてロボットの動作生成を行うため, Haptic Deviceを用いた人間の教示データを収集する.
  しかし, この際奥行き方向の情報の取得が難しいため, 鉗子をスピーディーにうまく操作することは困難である.
  そこで, 鉗子をゆっくりと正確に動かした際の軌跡から, 動作に対する制約を抽出し, これを組み込むことで教示データの質や模倣学習の精度を向上させることを行う.
  その手順は以下のようになっている.
  \begin{enumerate}
    \renewcommand{\labelenumi}{\arabic{enumi})}
    \item フェーズ遷移の条件記述
    \item 完璧なデモンストレーションからの動作制約抽出
    \item 制約に基づく力フィードバックを加えた教示によるデータ収集
    \item 得られたデータに基づく模倣学習
  \end{enumerate}

  1)それぞれの動作における制約を抽出するにあたり, それら制約の切り替わりとなるフェーズ遷移が起こる条件$S$を記述する.
  一例として, 鉗子をRubber Objectより少し上の位置に合わせ, その後鉗子を下に降ろし, 掴んで持ち上げるという動作を考える.
  このとき, 鉗子をRubber Objectの少し上の位置に合わせるには, xy平面方向に位置を合わせた上で, zについて最小値の制約を加えることができる.
  また, その後鉗子を下に下ろす動作では, 位置はそのままにzの最小値が下がり, Rubber Objectが把持可能になる.
  このときの$z^{ref}_{forcep}$の遷移を\figref{figure:fls-learning}に示す.
  この動きにおいて, フェーズの遷移の条件は, 例えばグリッパの開閉, または動作の収束とすることができる.
  つまり, $\bm{x}$の速度が小さい期間が続いた場合は位置合わせが収束したと判断し次に移る, また, 把持動作が起きたら掴んで持ち上げる動作が可能と判断し次に移る, といった形である.
  この他にも, 2つの鉗子の間の距離や速度の反転, 別の何らかのボタン等, 任意の形で制約を記述して良い.

  2)完璧なデモンストレーションが得られたとして, それが1)の条件$S$によって$N_{C}$個に区切られる.
  フェーズ$i$におけるデモンストレーションから, フェーズ$i$における動作制約$C_{i}$を抽出する($1\leq{i}\leqN_{C}$).
  1)と同じ例について考える(\figref{figure:fls-learning}).
  この場合, 最大値・最小値制約が効果的である.
  \figref{figure:fls-learning}のように$z_{\{0, 1, 2\}}$を用意すると, それぞれのフェーズにおける$z^{ref}_{forcep}$の制約は以下のように記述できる.
  \begin{align}
    C_1: z_1 < z^{ref}_{forcep} < z_2 \label{eq:constrain-example}\\
    C_2: z_0 < z^{ref}_{forcep} < z_1 \\
    C_3: z_0 < z^{ref}_{forcep} < z_2 
  \end{align}
  もちろん$x$や$y$について制約を課すこともできるが, 奥行きの認識が難しいことを考えると, $z$への制約が最も適切である.
  この他, 2つの鉗子の間の距離や速度等, 任意の制約を抽出することが可能である.

  3)ここでは2)で得られた制約を, 操作デバイスへの力フィードバックとして動作に組み込む.
  例えば\equref{eq:constrain-example}のような形の制約が得られた場合, これに基づく力フィードバックは以下のように記述できる.
  \begin{align}
    F_{z} =
    \begin{cases}
      k_{p}(z_1-z^{ref}_{forcep}) & \text{if $z^{ref}_{forcep} < z_1$}\\
      k_{p}(z_2-z^{ref}_{forcep}) & \text{if $z^{ref}_{forcep} > z_2$}
    \end{cases}
  \end{align}
  ここで, $F_{z}$は操作デバイスの$z$方向に対するフィードバック力, $k_{p}$は比例定数を指す.
  これは, $z$が$z_1$と$z_2$による制約の範囲から逸脱することを防ぐようなフィードバックをかけることに相当する.
  異なる制約に対しても, 同様に$F_x$, $F_y$, $F_z$に対する力フィードバックにより制約を組み込むことが可能である.
  この制約を加えたうえで教示によりデータを収集する.

  4)得られたデータに基づき模倣学習\cite{kawaharazuka2021imitation}を行う.
  人間の教示データには多様な動作速度や癖があるため, これらの情報をparametric bias \cite{tani2002parametric, ogata2005extracting}という学習可能なネットワーク入力変数により考慮する.
  以下のような予測モデル型のネットワークを学習させる.
  \begin{align}
    (\bm{s}_{t+1}, \bm{u}_{t+1}) = \bm{f}(\bm{s}_{t}, \bm{u}_{t}, \bm{p}) \label{eq:rnnpb}
  \end{align}
  ここで, $t$は現在のタイムステップ, $\bm{s}$はセンサ状態, $\bm{u}$は制御入力, $\bm{p}$はparametric bias, $\bm{f}$は予測モデルを表す.
  本研究で$\bm{s}$は, 画像情報をAutoEncoder \cite{hinton2006reducing}を通して次元圧縮した$\bm{\xi}$ ($\in \mathcal{R}^{12}$)を用いる.
  また, $\bm{u}$には左右の鉗子先端の指令値$\bm{x}^{ref}_{forcep-\{left, right\}}$ ($\in \mathcal{R}^{6}$)と, 鉗子の開閉である$h_{\{left, right\}}$ ($\in \{0, 1\}^{2}$)を用いる.
  これらのデータは10Hzで取得するものとする.

  一回のデモンストレーション$k$について, データ$D_k=\{(\bm{s}_1, \bm{u}_{1}), (\bm{s}_2, \bm{u}_2), \cdots, (\bm{s}_{T_{k}}, \bm{u}_{T_{k}})\}$を得る($1 \leq k \leq K$, $K$は全試行回数, $T_{k}$はその試行$k$に関する動作ステップ数とする).
  そして, 学習に用いるデータ$D_{train}=\{(D_1, \bm{p}_1), (D_2, \bm{p}_2), \cdots, (D_{K}, \bm{p}_K)\}$を得る.
  $\bm{p}_k$はその試行$k$に関するparametric biasであり, 人間の動作スタイルがこの空間に自己組織化される.
  データ$D_{train}$を用いて, 通常の模倣学習のようにネットワークの重み$W$のみを更新するのではなく, 同時に$\bm{p}_{k}$も更新していくことでRecurrent Neural Network with Parametric Bias (RNNPB)を学習させる.
  なお, 学習の際の損失関数はmean squared errorであり, 全$\bm{p}_k$は初期値を0として最適化される.
  タスクの実行方法はシンプルであり, 現在のセンサ状態$\bm{s}_{t}$と制御入力$\bm{u}_{t}$を取得し, \equref{eq:rnnpb}を順伝播することで$\bm{u}_{t+1}$を得て, これを実機に送ることを繰り返す.
  また, $h$については0.5を超えれば鉗子を閉じ, 0.5を下回れば鉗子を開く.
  なお, タスク実行時の$\bm{p}$は任意に設定でき, motion styleを変化させることができるが, 本研究では1)-3)の有無に関する比較がメインのため, $\bm{p}=\bm{0}$としている.
  % この際にも, 3)で得られた制約を課すことが可能である.
  % 例えば制約が\equref{eq:constrain-example}であるとき, 鉗子先端の指令値のz方向の値$z^{ref}_{forcep}$に最小最大値制約を加えながら動作することができる.

  なお, RNNPBは10層とし, 順に4層のfully-connected layer, 2層のLSTM layer \cite{hochreiter1997lstm}, 4層のfully-connected layerからなる.
  ユニット数については, \{$N_u+N_s+N_p$, 500, 300, 100, 100 (LSTMのunit数), 100 (LSTMのunit数), 100, 300, 500, $N_u+N_s$\}とした(なお, $N_{\{u, s, p\}}$は$\{\bm{u}, \bm{s}, \bm{p}\}$の次元数とする).
  activation functionはhyperbolic tangent, 更新則はAdam \cite{kingma2015adam}とした.
  画像を圧縮する際は, $128\times96$のRGB画像について, kernel sizeが3, strideが2の畳み込み層を5回適用し, 全結合層で順にユニット数1024, 12まで次元を削減したあと, 同様に全結合層・逆畳み込み層によって画像を復元していく形を取っている.
  最終層以外についてはbatch normalization \cite{ioffe2015batchnorm}が適用され, activation functionは最終層以外についてはReLU \cite{nair2010relu}, 最終層はSigmoid, 更新則はAdam \cite{kingma2015adam}とした.
  学習の際にはデータに多様なランダムノイズを入れる\cite{florence2020imitation}.
  $\begin{pmatrix}\bm{s}^{T}&\bm{u}^{T}\end{pmatrix}^{T}$について, 前ステップとの差を計算し, その共分散$\bm{\Sigma}$を得る.
  平均を$\bm{0}$, 共分散を$0.3^{2}\Sigma$とした多変量正規分布に従う乱数を毎ステップ加えたうえでデータを学習に利用する.
}%

\begin{figure}[t]
  \centering
  \includegraphics[width=0.95\columnwidth]{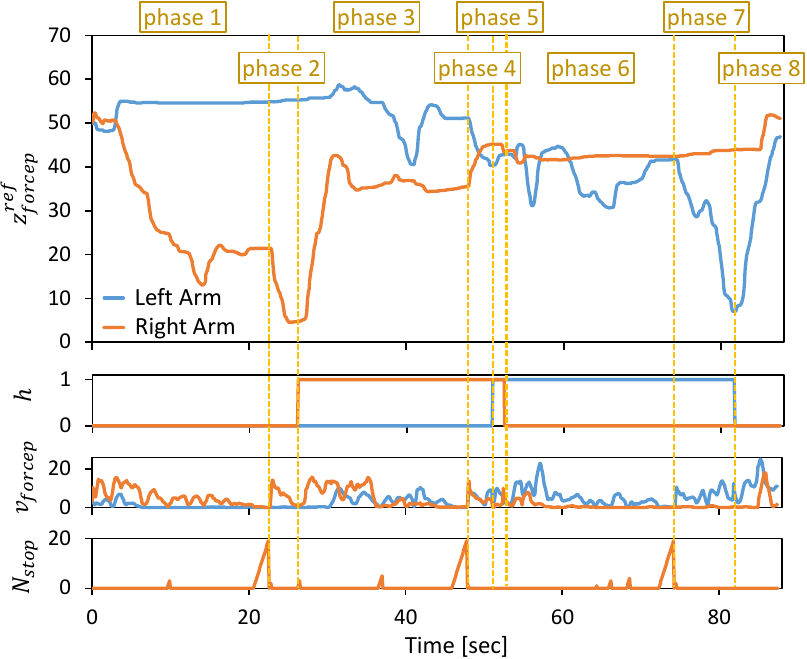}
  \vspace{-1.0ex}
  \caption{The transition of $z^{ref}_{forcep}$, $h$, and $v_{forcep}$ of the left and right arms, and the transition of $N_{stop}$ for constraint generation from one exemplary demonstration of peg transfer.}
  \vspace{-1.0ex}
  \label{figure:complete-graph}
\end{figure}

\begin{figure}[t]
  \centering
  \includegraphics[width=0.95\columnwidth]{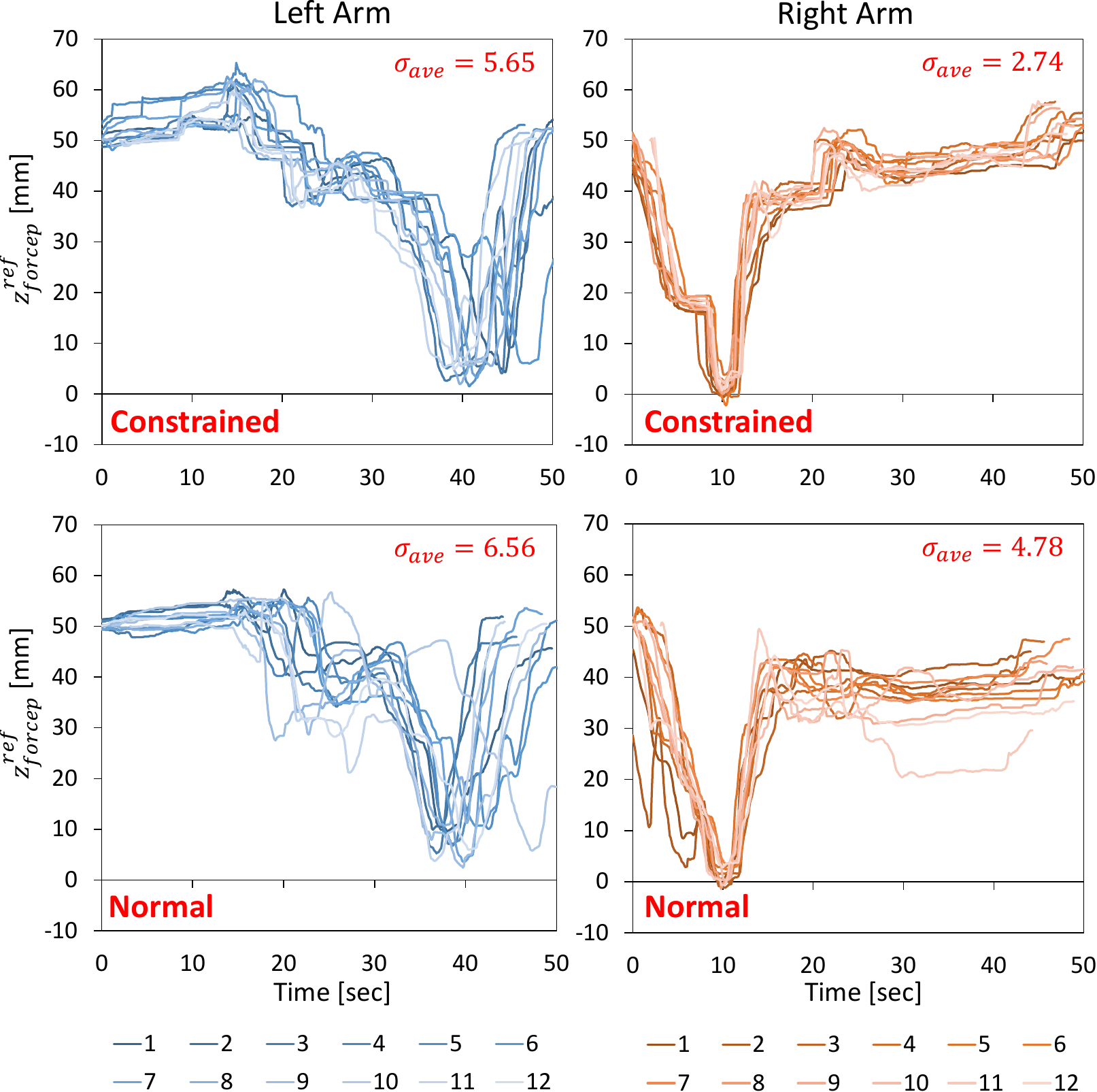}
  \vspace{-1.0ex}
  \caption{The transition of $z^{ref}_{forcep}$ of 12 demonstrations of peg transfer with constrained or normal data collection for imitation learning.}
  \vspace{-3.0ex}
  \label{figure:collect-graph}
\end{figure}

\section{Experiments} \label{sec:experiment}

\subsection{Data Collection} \label{subsec:collect-exp}
\switchlanguage%
{%
  First, we make a single exemplary demonstration.
  In this study, we describe the transition condition $S$ and the motion constraint $C_{i}$ in the same way as in the example of \secref{subsec:fls-learning} (\figref{figure:fls-learning}).
  For $S$, when the state where L2 norm of the forcep-tip velocity $v_{forcep}=||\dot{\bm{x}}_{forcep}||_{2}$ is less than 1.0 regarding both right and left forceps continues for $N^{thre}_{stop}$ steps, or when the opening/closing state $h$ of either left or right forceps changes, the phase is transitioned.
  Let $N_{stop}$ be the number of consecutive steps in the state where $v_{forcep}<1.0$ regarding both forceps.
  We set $N^{thre}_{stop}=20$ when generating constraints and $N^{thre}_{stop}=10$ when actually collecting data with the generated constraints.
  When generating the constraints, the motion itself is not used for learning, so it does not matter how slowly the robot is moved for the single exemplary demonstration.
  The robot should be moved carefully in the $z$ direction in particular.
  For $C_{i}$, the maximum and minimum values of $z^{ref}_{forcep}$ for the left and right forceps are obtained from the start and end points of the motion in phase $i$.
  Since the minimum and maximum values are calculated from the start and end points of the motion phase, we only need to be aware of the phase transition even if the overall motion is unsteady.
  The phase transition can be viewed visually on the monitor.
  The demonstration is shown in \figref{figure:complete-exp}.
  The transition of $z^{ref}_{forcep}$, $h$, and $v_{forcep}$ of the left and right forceps, and the transition of $N_{stop}$ are shown in \figref{figure:complete-graph}.
  First, we approach the peg in which the rubber object is stuck, aim at the top of the rubber object, grasp it by lowering the right forceps, and lift it up.
  Next, the left forceps is aimed at the top of the rubber object when it is at the center of the peg board, the left forceps is lowered to grasp it, and the right forceps releases the rubber object.
  Finally, we aim at the top of the left peg, lower the left forceps and put the rubber object into the peg, then release it and return to the original position.
  $N_{stop}$ increases when $v_{forcep}$ of the left and right forceps approaches 0, and the phase shifts when $N_{stop}$ reaches $N^{thre}_{stop}$.
  The phase is also shifted each time $h$ changes.
  In the end, $N_{C}=8$ phases were generated.
  From the motion at each phase, the constraint of $z^{ref}_{forcep}$ and the force feedback law based on the constraint are automatically generated.

  Next, we collected data for actual imitation learning.
  The robot performed the action of transferring a rubber object from the right peg to the left peg, for each of the three pegs on the right side, four times each, totaling 12 trials.
  During this process, data was collected for both constrained data collection (Constrained), where the action was constrained by force feedback, and normal data collection (Normal) without any constraints.
  The trajectories of $z^{ref}_{forcep-\{left, right\}}$ for the 12 trials are shown in \figref{figure:collect-graph}.
  % The data is synchronized based on the grasping time of the rubber object.
  For both hands, Constrained shows more stable and consistently close values for $z^{ref}_{forcep}$ compared to Normal in each trial.
  The average variance $\sigma_{ave}$ for Constrained is 5.65 for the left arm and 2.74 for the right arm, whereas for Normal, it is 6.56 for the left arm and 4.78 for the right arm.
  Imposing minimum and maximum constraints on the $z$-directional movement allows for stable data collection.
}%
{%
  まずは完璧なデモンストレーションを一回成功させる.
  本研究では\secref{subsec:fls-learning}の例(\figref{figure:fls-learning})と同様の形で, 遷移条件$S$と動作制約$C_{i}$を記述する.
  $S$については, 左右両方の鉗子先端位置$\bm{x}_{forcep}$の速度のL2ノルム$v_{forcep}=||\dot{\bm{x}}_{forcep}||_{2}$が1.0以下である状態が$N^{thre}_{stop}$ステップ以上続いた場合, または左右どちらかの鉗子の開閉$h$が変化した場合にフェーズを遷移させる.
  なお, $v_{forcep}<1.0$である状態の連続するステップ数を$N_{stop}$とする.
  また, 制約を得る際は$N^{thre}_{stop}=20$, 実際に制約付きでデータ収集を行う際は$N^{thre}_{stop}=10$とした.
  制約を得る際は, その動作自体を学習に利用するわけではないためどんなにゆっくりと動かしても良く, 特に$z$方向には丁寧に動かすべきである.
  $C_{i}$については, 左右それぞれの鉗子の$z^{ref}_{forcep}$の最大値と最小値を, フェーズ$i$の動作の始点と終点から取得する.
  なお, 動作の始点と終点から最小値と最大値が計算されるため, 行ったり戻ったりふらふらした動作でも, フェーズの切り替わりのみ意識すれば問題ない.
  フェーズの切り替わりはモニタから視覚的に得ることができる.
  デモンストレーションの様子を\figref{figure:complete-exp}に, デモンストレーションから得られた左右の鉗子の$z^{ref}_{forcep}$, $h$, $v_{forcep}$と, $N_{stop}$の遷移を\figref{figure:complete-graph}に示す.
  まずはrubber objectの刺さったpegに向かい, rubber objectの上部で狙いを定め, 鉗子を降ろしてそれを把持する.
  次に視野中心で左の鉗子がrubber objectの上部に来るよう狙いを定め, 鉗子を降ろしてそれを把持し, 右の鉗子をrubber objectから離す.
  左のpegの上部で狙いを定め, 鉗子を降ろしてrubber objectをpegに刺し, それを離して元の位置に戻る.
  左右の鉗子の$v_{forcep}$が0に近づくと$N_{stop}$が上昇し, これが$N^{thre}_{stop}$に到達した時点でフェーズが移行する.
  また, $h$が変化する度にもフェーズが移行し, 最終的に$N_{C}=8$のフェーズが生成された.
  ここから自動的に$z^{ref}_{forcep}$の制約, またそれに基づく力フィードバック則が生成される.

  次に, 実際に模倣学習に利用するデータを収集した.
  ロボットから見て右側のペグに入ったrubber objectを左のペグに移す動作を, 右側の3つのペグそれぞれについて4回ずつ, 計12回行った.
  この際,力フィードバックにより制約を受けた教示(Constrained)と, 通常の一切制約の無い教示(Normal)によりそれぞれデータを収集している.
  この際の$z^{ref}_{forcep-\{left, right\}}$の12回分の軌跡を\figref{figure:collect-graph}に示す.
  なお, rubber objectの把持の時間でデータを揃えている.
  どちらの手についても, NormalよりもConstrainedの方が$z^{ref}_{forcep}$が毎回近い値で安定している.
  実際, その分散の平均$\sigma_{ave}$はConstrainedの場合左腕が5.65, 右腕が2.74なのに対して, Normalの場合左腕が6.56, 右腕が4.78となっている.
  $z$方向の動きに最小値・最大値制約を課すことで, 安定したデータ収集が可能である.
}%

\begin{figure*}[t]
  \centering
  \includegraphics[width=1.9\columnwidth]{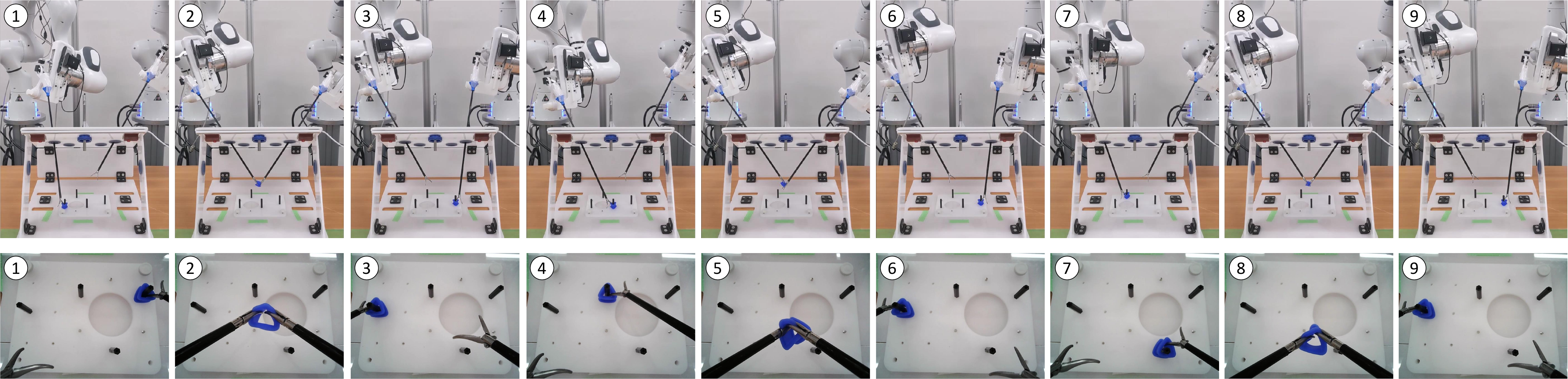}
  \vspace{-1.0ex}
  \caption{Peg transfer experiment using imitation learning trained by data obtained with constrained data collection.}
  \vspace{-3.0ex}
  \label{figure:task-exp}
\end{figure*}

\begin{figure}[t]
  \centering
  \includegraphics[width=0.85\columnwidth]{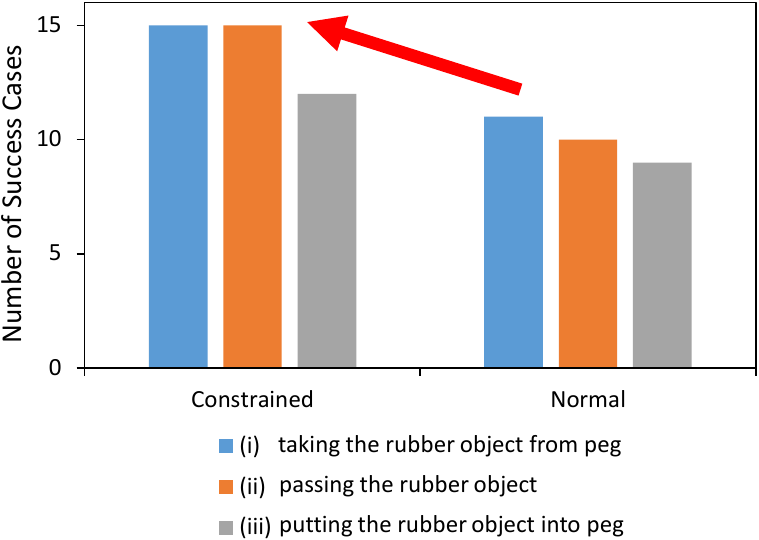}
  \vspace{-1.0ex}
  \caption{Comparison of success rates of peg transfer using imitation learning trained by data obtained with constrained or normal data collection.}
  \vspace{-1.0ex}
  \label{figure:task-graph}
\end{figure}

\begin{figure}[t]
  \centering
  \includegraphics[width=1.0\columnwidth]{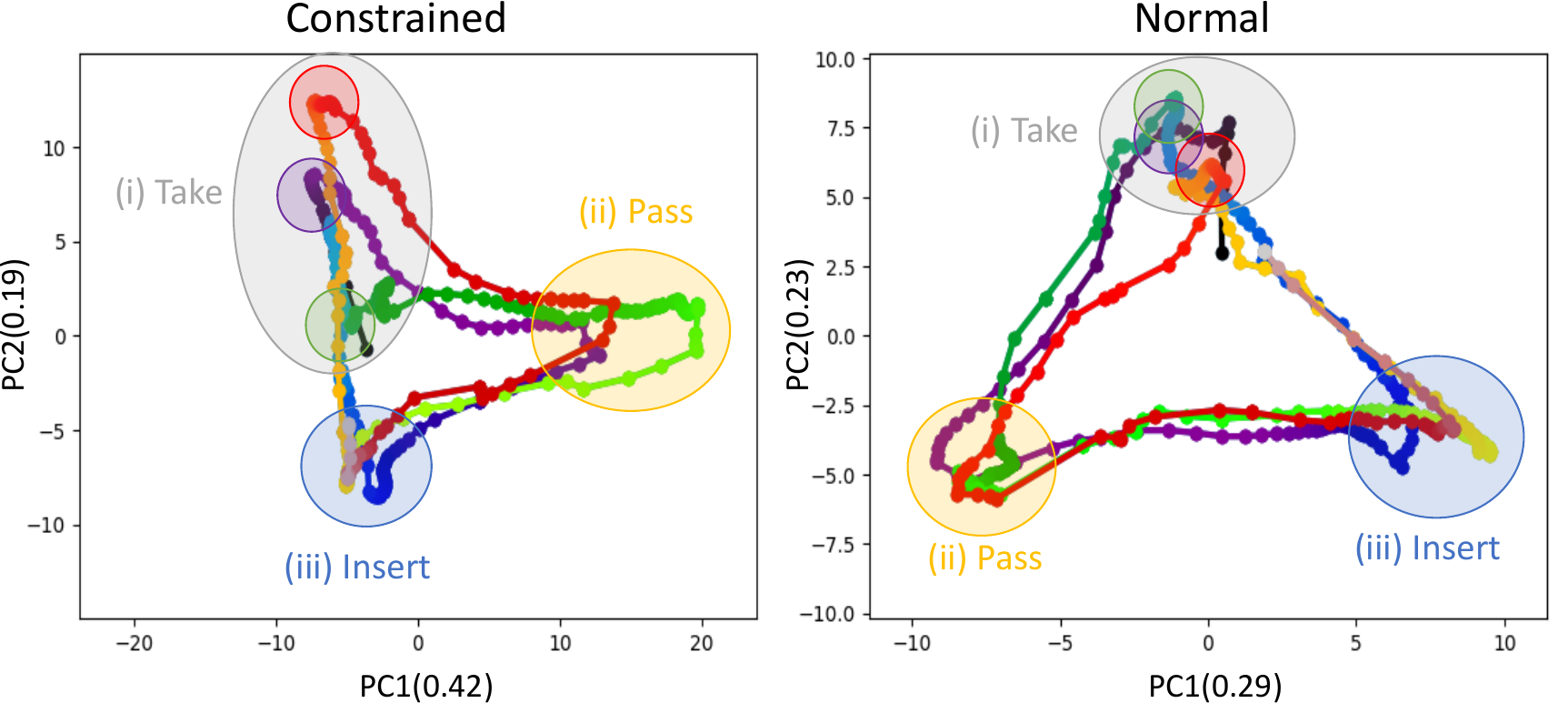}
  \vspace{-3.0ex}
  \caption{Comparison of the trajectories of LSTM's latent space when performing peg transfer using imitation learning with constrained or normal data collection. Each graph shows successive peg transfer for three pegs, and the color changes gradually with time.}
  \vspace{-3.0ex}
  \label{figure:task-pca}
\end{figure}

\subsection{Task Execution with Imitation Learning} \label{subsec:task-exp}
\switchlanguage%
{%
  Imitation learning was performed using the 12 sets of data collected in \secref{subsec:collect-exp}.
  The actual task execution using imitation learning with constrained data collection is shown in \figref{figure:task-exp}.
  Even when the peg that the rubber object is put on is changed in turn, the robot is able to grasp, deliver, and insert the rubber object appropriately.
  Although the position and the way of holding the rubber object are different in each case, the delivery of the rubber object is successful in all cases.

  Next, we compared the imitation learnings based on Constrained and Normal data collection.
  We conducted a total of 15 experiments, in which the rubber object in each peg was transferred 5 times each.
  The number of successful cases of (i) taking the rubber object from a peg, (ii) passing the rubber object from the right forceps to the left, and (iii) inserting the rubber object into a peg are shown in \figref{figure:task-graph}.
  In imitation learning based on Constrained data collection, all operations except for the last insertion have succeeded, and the success rate is very high.
  % Note that (iii) is the most difficult operation because the grasping posture of the rubber object is very different each time.
  On the other hand, in imitation learning based on Normal data collection, there were many cases where the peg was not grasped properly at (i), and the success rate is lower compared to using Constrained data collection.
  Although the peg's position remains consistent between data collection and this experiment, it is found that the success rate of imitation learning varies significantly depending on the presence or absence of the Constrained data collection.

  In addition, for the data of imitation learning based on Constrained and Normal data collection, where the peg that the rubber object is put on is changed in turn, we applied Principle Component Analysis (PCA) to LSTM's latent space and visualized the results in two dimensions in \figref{figure:task-pca}.
  In both cases, we observed that the latent space evolves in the order of (i) take, (ii) pass, and (iii) insert.
  Regarding (i), with Constrained data collection, the trajectories of the latent space are distinct for each of the three pegs.
  However, with Normal data collection, they closely follow the same trajectory.
  Also, (ii) should exhibit variations in movements depending on how the rubber object is held and where it is passed to, and (iii) should mostly result in similar movements since there is only one peg available.
  These characteristics are more evident with Constrained data collection, while they are less pronounced with Normal data collection.
}%
{%
  \secref{subsec:collect-exp}で得られた12回分のデータを用いて模倣学習を行った.
  Constrainedなデータ収集を用いた模倣学習における実際のタスクの様子を\figref{figure:task-exp}に示す.
  rubber objectの刺さるペグの位置を順に変化させても, 適切に把持, 受け渡し, 挿入を行うことができている.
  受け渡しも毎回位置や持ち方が異なるが, 正しく成功していることがわかる.

  次に, ConstrainedとNormalなデータ収集に基づく模倣学習を比較した.
  それぞれのペグに入ったrubber objectを移す動作を5回ずつ, 計15回の実験を行った.
  この際の(i)rubber objectをペグから取り出す動作, (ii) rubber objectを右手から左手に受け渡す動作, (iii) rubber objectをpegに挿入する動作, の成功回数をそれぞれ\figref{figure:task-graph}に示す.
  Constrainedの場合は最後の挿入以外では全て成功しており, 成功率は非常に高い.
  % なお, (iii)の挿入部は, 毎回ペグの把持姿勢が大きく異なるため, 最も難しい動作となっている.
  一方で, Normalの場合は(i)の把持の時点で上手く行かない場合も多く, Constrainedに比べてその成功率が落ちていることがわかる.
  ペグの位置はデータ収集時と本実験で同じであるものの, 制約付きデータ収集の有無によって, 模倣学習の成功率が大きく変化することがわかる.

  また, ConstrainedとNormal data collectionについて, 各ペグに入ったrubber objectを一回ずつ移動させる動きを行い, その際のLSTMの潜在空間についてPCAをかけて2次元に描画した際の結果を\firef{figure:task-pca}に示す.
  両者ともに, (i) pick, (ii) pass, (iii) insertの順に潜在空間が変化していることがわかる.
  その一方で, (i) pickについて, Constrainedでは3つのペグそれぞれに対する潜在空間の軌跡が異なっているのに対して, Normalではほぼ同じ軌跡を辿ってしまっている.
  また, (ii)はrubber objectの持ち方や受け渡しの場所に応じて動きが異なり, (iii)はペグが一つしかないためほとんど同じ動きになるはずである.
  Constrainedではその特性が良く読み取れる一方, Normalについてはその特性は強く読み取れない.
}%

\section{Discussion} \label{sec:discussion}
\switchlanguage%
{%
  In this study, we have extracted phase-specific motion constraints from a single exemplary demonstration, and constructed force feedback based on the constraints to overcome the difficulty of motion in the depth direction, especially for a monocular camera.
  The introduction of motion constraints enables more stable data collection.
  Imitation learning with the constrained data collection is more stable than imitation learning with the unconstrained data collection, and is able to realize the peg transfer task with higher accuracy.

  We discuss the limitations and future prospects of this study.
  First, it is necessary for humans to appropriately determine which variables to examine for the phase transition and which constraints to generate for each variable.
  Although this idea itself can be used for various tasks, we believe that the scope of application will be expanded if the system can autonomously determine the phase transition and the form of constraints depending on the task.
  This is a similar idea to the discovery of the knacks \cite{kuniyoshi2004knack}.
  In the future, we would like to develop a robot that can grow autonomously.
  % In addition, this study is not yet ready for actual use in the real world.
  Second, in reality, the approach to the inside of the stomach, which is a place with considerable individual differences and gradual changes, requires a more adaptive robot system.
  Since the current setup is too constrained to operate inside of the stomach, for example, it would be better to apply a constraint only on the distance between the two forceps.
  There is also the issue of whether the constraints can be determined from a single exemplary demonstration, and further study is needed.
  Our idea is very simple but effective and can also be used in other fields outside surgery in the future.
}%
{%
  得られた結果についてまとめる.
  本研究では一度の模範となる動作からフェーズごとの動作制約を抽出, 力フィードバックを組むことで, 特に単眼における奥行き方向への動作の難しさを克服した.
  動作制約を導入することで, より安定したデータ収集が可能になる.
  制約付きで得られたデータを用いた模倣学習は, 制約を入れずに得られたデータを用いた模倣学習に比べ動作が安定しており, 高い精度でPeg Transferタスクを実現することが可能であった.

  本研究の限界と今後の展望について述べる.
  まず, 現状はどの変数を見てフェーズ移行するかや, どの変数に対して制約を生成するかは人間が適切に決定する必要がある.
  本アイデア自体は多様なタスクに利用可能であるが, タスクに依存したフェーズ移行や制約の形を自律的に決定することができればより適用範囲が広がると考える.
  これはコツの発見と似た考え方である\cite{kuniyoshi2004knack}.
  今後, 体表ポートの制約や力フィードバックの生成も自動生成するといった, 自律的に成長するロボットの開発を進めたい.
  次に, 本研究はまだ実際に現場で使えるとは言い難い.
  実際にはお腹の中というかなり個人差や変化の激しい場所に対してアプローチするため, より適応的なシステムが求められる.
  現状のセットアップでは制約が厳しいため, 二つの鉗子の距離に対してのみ受け渡し制約をかける等も可能である.
  また, たった一回の模範動作から制約を決定していいのかという問題もあり, 今後のさらなる検討が必要である.
  本研究のアイデアは非常にシンプルではあるが効果的であり, 手術以外の分野にも今後利用可能であると考えている.
}%

\section{CONCLUSION} \label{sec:conclusion}
\switchlanguage%
{%
  In this study, we proposed a method to perform the peg transfer task in Fundamentals of Laparoscopic Surgery (FLS) as a setup toward robotic laparoscopic surgery.
  We developed an overall system which includes two robots with forceps, two haptic devices, and control architecture based on constrained inverse kinematics and constrained imitation learning.
  In particular, constrained imitation learning allows for easy data collection and precise control using only monocular images, without the need for depth images or models of a target to be operated on.
  By first generating minimum and maximum constraints for the forceps depth from single exemplary demonstration and collecting data using these constraints for imitation learning, it is possible to easily perform the peg transfer task with higher accuracy.
  The idea of automatic extraction of motion constraints can be utilized for various robots, tasks, and environments.
  % In the future, we aim to advance the development of more practical systems for robotic laparoscopic surgery.
}%
{%
  本研究ではロボットによる腹腔鏡手術に向け, FLSにおけるペグ移動を行う戦略について紹介した.
  鉗子を持った2台のロボットと2台のハプティックデバイス, 制約付き逆運動学と制約付き模倣学習に基づく制御アーキテクチャを含めた全体システムを構築した.
  特に制約付き模倣学習は深度画像や対象のモデルを用いることなく, 単眼画像のみで正確なデータ収集と制御を行うことが可能である.
  最初に模範となる動作から奥行き方向に対する最小値最大値制約を生成し, これを用いてデータ収集と模倣学習を行うことで, 容易に高い精度でペグ移動を行うことが可能になった.
  本研究の自動制約抽出の考え方は多様なタスクや環境に利用可能である.
  今後, より実用的なシステムの構築や, 他のタスクへの適用を進めたい.
}%

\section*{Acknowledgement}
This work was partially supported by JST Moonshot R\&D under Grant Number JPMJMS2033.

{
  %\footnotesize
  %\small
  %\bibliographystyle{junsrt}
  \bibliographystyle{IEEEtran}
  \bibliography{main}
}

\end{document}